\documentclass[10pt,twocolumn,letterpaper]{article}

\usepackage{cvpr}              

\usepackage[dvipsnames]{xcolor}

\definecolor{cvprblue}{rgb}{0.21,0.49,0.74}
\usepackage[pagebackref,breaklinks,colorlinks,citecolor=cvprblue]{hyperref}

\usepackage{graphicx}
\usepackage{amsmath}
\usepackage{amssymb}
\usepackage{booktabs}
\usepackage{multirow}
\usepackage{subcaption}
\usepackage{makecell}
\usepackage[accsupp]{axessibility} 

\usepackage[capitalize]{cleveref}
\crefname{section}{Sec.}{Secs.}
\Crefname{section}{Section}{Sections}
\Crefname{table}{Table}{Tables}
\crefname{table}{Tab.}{Tabs.}

\def\confName{CVPR}
\def\confYear{2024}

\begin{document}

\title{Generalizing 6-DoF Grasp Detection via Domain Prior Knowledge}

\author{Haoxiang Ma$^{1,2}$\quad Modi Shi$^{1,2}$\quad Boyang Gao$^{3,4}$\quad Di Huang$^{1,2}$\thanks{Corresponding author.}\\
$^{1}$State Key Laboratory of Software Development Environment, Beihang University, Beijing, China\\
$^{2}$School of Computer Science and Engineering, Beihang University, Beijing, China\\
$^{3}$School of Computer Science and Technology, Harbin Institute of Technology, Harbin, China\\
$^{4}$Geometry Robotics\\
{\tt\small \{mahaoxiang822,modishi,dhuang\}@buaa.edu.cn, \{boyang.gao\}@geometryrobot.com
}
}

\maketitle

\begin{abstract}
   We focus on the generalization ability of the 6-DoF grasp detection method in this paper. While learning-based grasp detection methods can predict grasp poses for unseen objects using the grasp distribution learned from the training set, they often exhibit a significant performance drop when encountering objects with diverse shapes and structures. To enhance the grasp detection methods' generalization ability, we incorporate domain prior knowledge of robotic grasping, enabling better adaptation to objects with significant shape and structure differences. More specifically, we employ the physical constraint regularization during the training phase to guide the model towards predicting grasps that comply with the physical rule on grasping. For the unstable grasp poses predicted on novel objects, we design a contact-score joint optimization using the projection contact map to refine these poses in cluttered scenarios. Extensive experiments conducted on the GraspNet-1billion benchmark demonstrate a substantial performance gain on the novel object set and the real-world grasping experiments also demonstrate the effectiveness of our generalizing 6-DoF grasp detection method. Code is available at \url{https://github.com/mahaoxiang822/Generalizing-Grasp}.
\end{abstract}

\section{Introduction}
\label{sec:intro}
\begin{figure}[t]
\centering
\begin{subfigure}[]{0.85\columnwidth}
\centering
\includegraphics[width=\textwidth]
{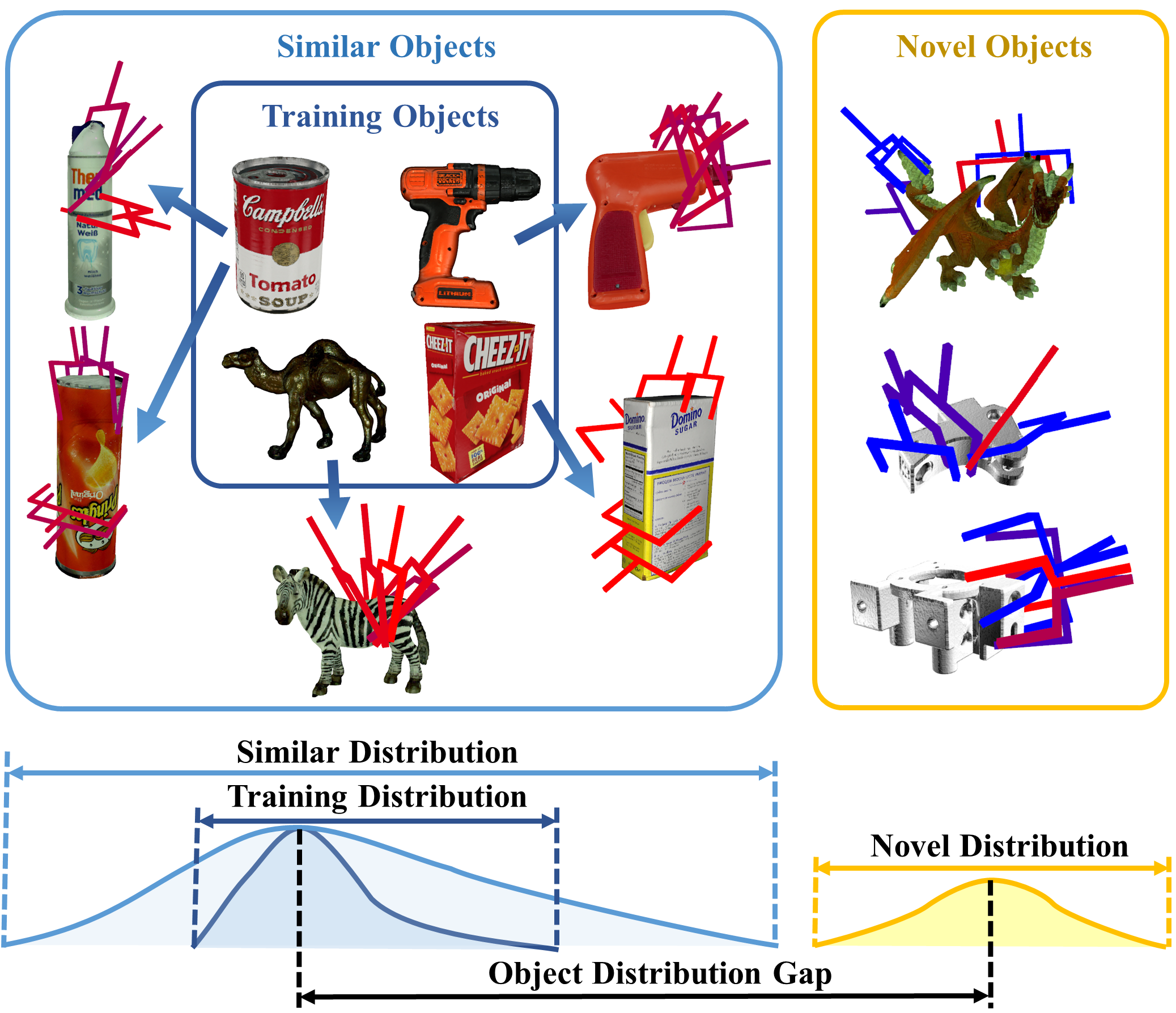}
\caption{}
\end{subfigure}
\begin{subfigure}[]{0.8\columnwidth}
\centering
\includegraphics[width=\textwidth]{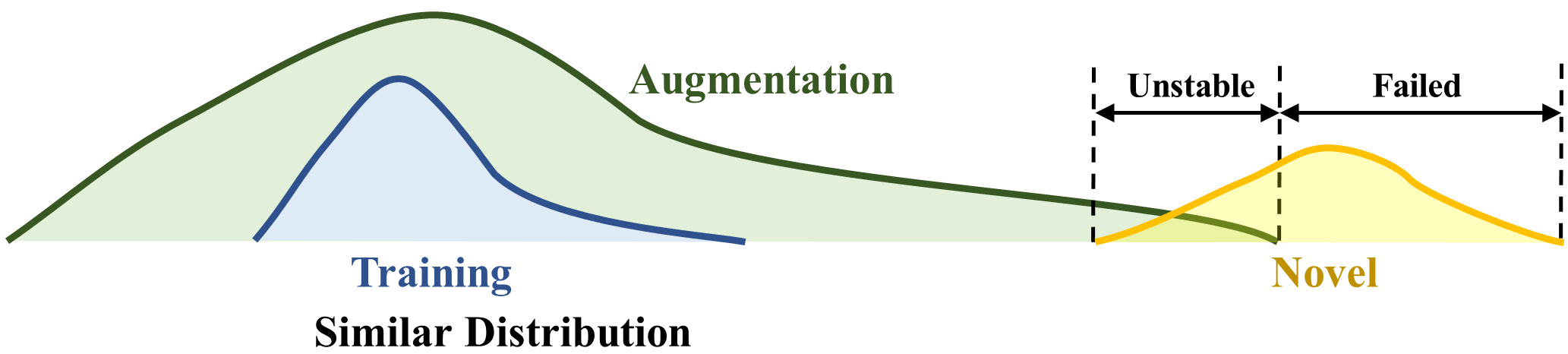}
\caption{}
\end{subfigure}
\caption{Illustration of (a) the performance and distribution gap between objects similar to training samples (\textbf{similar objects}) and those with largely varied shapes and structures (\textbf{novel objects}) and (b) the limitations of the object augmentation methods.}
\label{fig:1}
\vspace{-0.5cm}
\end{figure}

Given an object, robotic grasp detection aims to find suitable and sufficient gripper configurations for various manipulation tasks. Traditional methods \cite{chen1993finding,pollard2004closure,roa2009computation} establish hand-crafted criteria to evaluate grasp samples according to 3D models of objects. Despite providing precise interaction between objects and grippers, they assume that object models are available in advance and also suffer a slow running speed, which limits their popularization. As the development of deep learning, data-driven methods \cite{DBLP:conf/rss/MahlerLNLDLOG17, DBLP:journals/ijrr/PasGSP17, DBLP:conf/iros/DepierreD018} have been widely studied. These methods predict grasps without the need for pre-prepared 3D models and demonstrate the capability to handle unseen objects. However, they often struggle when encountering objects whose shapes and structures significantly deviate from those in the training set, making them difficult to adapt to more diverse applications. As illustrated in Fig. \ref{fig:1} (a), learning-based grasp detection methods are basically able to transfer to objects similar to those appearing in training, but they do not effectively generalize to novel objects with their distribution greatly changed.

To facilitate the generalization of learning-based grasp detection methods towards a variety of unseen objects, previous attempts usually apply data augmentation techniques to expand the distribution of the training set. More objects are typically produced by randomly concatenating a number of 3D primitives \cite{tobin2018domain,bousmalis2018using} or by directly synthesizing through generative networks \cite{morrison2020egad, jiang2022learning}. They do enrich the diversity of objects for training, which leads to a gain in overall performance. However, since such augmentation is conducted only according to original data, it contributes more to similar objects whose distribution is largely overlapped and offers little help to new objects whose distribution is not well covered. As depicted in Fig. \ref{fig:1} (b), on the one hand, to the objects that are far from the center of training samples in the feature space, totally wrong grasps could be delivered. On the other hand, for the ones that are near the center of training samples in the feature space, although reasonable grasps may be obtained, the predictions tend to be unstable due to the distribution gap. Both cases make those methods problematic to generalize to unseen targets.

To address the dilemma above, inspired by informed machine learning \cite{karniadakis2021physics,von2021informed}, we introduce domain prior knowledge of robotic grasping. Compared to data augmentation, domain prior knowledge does not depend on the distribution of training data, allowing for easy adaptation to objects with significant shape and structure differences. Firstly, to enable the grasp detection network to generalize to novel objects, we incorporate certain physical rules on grasping, such as force conditions \cite{DBLP:journals/ijrr/Nguyen88} and contact positions, which are typically used in the traditional analytical grasping methods. The physical rules provide valuable clues in terms of grasp stability regardless of object-specific properties, thus motivating us to integrate such rule-based priors into the grasp detection network. Secondly, to deal with the unstable results predicted by the network, we introduce another type of knowledge about the interaction of the gripper and object. The contact map \cite{DBLP:conf/cvpr/BrahmbhattHKH19} is able to denote the regions correlated with grasping as well as the kinematics of the gripper. By employing a neural network to encode the preferred object regions for optimal grasps, the learning-based contact map prior can be utilized to refine insufficiently accurate grasps.

Concretely, in this paper, we propose a generalized 6-DoF grasp detection framework with domain prior knowledge, which consists of two components: Physical Constraint Regularization (\textbf{PCR}) and Contact-Score Joint Optimization (\textbf{C-SJO}). In PCR, we integrate the physical prior into the network as the regularization, thereby constraining the correlation between gripper poses and object models. To enable the back-propagation of PCR, we employ an end-to-end 6-DoF grasp detection network for grasp prediction and utilize the Signed Distance Field (SDF) to encode the object model, facilitating the differentiable computation of physical constraints. Compared to fitting grasp annotations directly, PCR guides the network to predict grasps following object-independent physical rules, thereby enhancing the generalization capability for novel objects. To refine the unstable prediction for novel objects, we introduce the C-SJO at test time based on the contact map prior. The contact map represents the contact region on the object's surface by calculating the distance between the gripper model and the objects. A contact map prediction network is introduced to encode the contact pattern of good grasps by learning from grasp labels. By aligning the contact map of the current grasp and the prediction from the network, the grasp pose can be optimized within its neighborhood. However, as for 6-DoF grasp detection in clutter, using the euclidean distance for contact map calculation cannot adequately address inaccurate contact positions. Therefore, we introduce a projection contact map to solve this problem. Additionally, due to the noise from the depth sensor and the occlusion in cluttered scenarios, relying solely on the contact prior can lead to singular results. To mitigate this, we use score optimization to constrain the searching space of contact optimization, in which a grasp score network is utilized. 

The contribution of this paper can be summarized as:

\begin{itemize}
    \item We propose a domain prior knowledge informed 6-DoF grasp detection framework to enhance the generalization ability for novel objects.
    \item We design the physical constraint regularization to represent the physical prior for 6-DoF grasping and integrate it into the network in a differential strategy.
    \item We employ a contact-score joint optimization with the projection contact map, which is suitable for refining the inaccurate prediction in cluttered scenarios.
\end{itemize}


\section{Related work}

\noindent \textbf{6-DoF Grasp Detection} To generate diverse and feasible grasping in cluttered scenes, 6-DoF grasp detection has been advanced recently. Compared to planar grasp detection where the grasp space is limited \cite{DBLP:conf/rss/MorrisonLC18, DBLP:conf/iros/DepierreD018, DBLP:conf/icra/QinMGH23}, 6-DoF grasp detection method can predict grasping in $SE(3)$ space, thereby supporting more complex downstream task. \cite{DBLP:journals/ijrr/PasGSP17} pioneers a sampling and evaluation framework for 6-DoF grasp detection. Using point-cloud input, they heuristically sample grasp candidates and employ a Convolutional Neural Network (CNN) for scoring them. \cite{DBLP:conf/icra/LiangMLGTFS019} build upon the framework, utilizing a neural network for point-cloud data to achieve improved grasp sample evaluation. More recently, a series of studies \cite{DBLP:conf/iccv/MousavianEF19, DBLP:conf/corl/QinCZSXS19, DBLP:conf/corl/BreyerCOSN20, DBLP:conf/cvpr/FangWGL20, DBLP:conf/iccv/WangFGFGL21, DBLP:conf/corl/Ma022,DBLP:conf/rss/0002ZSFZ21, DBLP:conf/icra/DaiZGRZW23} propose end-to-end strategy for grasp prediction. \cite{DBLP:conf/iccv/MousavianEF19} employs a variational auto-encoder for grasp generation given the object point-clouds. \cite{DBLP:conf/corl/QinCZSXS19} introduces a single-shot grasp proposal network for efficient grasp detection based on single view point-clouds, and \cite{DBLP:conf/corl/BreyerCOSN20} employs a Truncated Signed Distance Function (TSDF) to map multiple frames and predict grasps in voxel space. \cite{DBLP:conf/cvpr/FangWGL20} provides a large benchmark with densely annotated grasp labels and designs a baseline method for dense grasp prediction. Subsequently, \cite{DBLP:conf/iccv/WangFGFGL21} defines graspness to represent grasp probability in searching space and \cite{DBLP:conf/corl/Ma022} focuses on the scale imbalance problem for 6-DoF grasp detection. With the development of neural representations, \cite{DBLP:conf/rss/0002ZSFZ21} utilizes an occupancy network to learn a shared representation between 3D reconstruction and grasp detection. \cite{DBLP:conf/icra/DaiZGRZW23} introduces a generalizable neural radiance field for the grasp detection of transparent and specular objects. Despite these advancements in 6-DoF grasp detection, the performance for novel objects, particularly those with varied shapes and structures, remains sub-optimal. In this paper, we investigate this shortcoming.

\noindent \textbf{Generalization on Grasp Detection} Several strategies are proposed to enhance the generalization capability of grasp detection, which mainly focuses on enriching the distribution of training objects. \cite{bousmalis2018using} procedurally generates objects by attaching rectangular prisms at random locations and orientations. To enhance the shape diversity, \cite{tobin2018domain} constructs a set of object primitives by decomposing everyday objects and generates diverse objects by randomly sampling these primitives and combing them. In addition to heuristically generating random objects, \cite{morrison2020egad} provides the Evolved Grasping Analysis Dataset (EGAD), which includes objects of varying shape complexity and grasp difficulty generated by the 3D compositional pattern producing networks. \cite{jiang2022learning} employs an AutoEncoder-Critic network to interpolate new shapes from two objects for augmentation. Besides augmenting current shapes to improve the performance of the learning-based grasp detection method, \cite{DBLP:conf/case/WangTLJGDMIG19} explores the task of generating adversarial objects that are difficult to grasp. Different from previous augmentation methods for generalized grasp detection, we introduce the grasp domain knowledge to our framework explicitly. Therefore, our method doesn't rely on the distribution of training objects, resulting in an improvement for out-of-distribution objects.

\noindent \textbf{Usage of domain knowledge in Grasping} Domain prior knowledge has been leveraged in various grasping applications. Early analytical methods \cite{DBLP:conf/icra/FerrariC92, DBLP:journals/ram/MillerA04, DBLP:journals/ijrr/RodriguezMF12} make use of the physical prior knowledge to examine the force dynamics between the gripper and the object. More recently, several methods incorporate some grasp prior knowledge into the hand-object grasp synthesis. \cite{DBLP:journals/ral/LiuLJZZ22} introduces a differentiable force closure algorithm designed to optimize hand configurations, which facilitates the fast generation of physically stable grasps. \cite{DBLP:conf/rss/LiuP0GM20} designs a generalized Q1 metric that serves as a loss for a grasp planner, producing precise multi-finger grasping for single object with watertight model or rendered depth images. While the aforementioned methods employ domain prior knowledge to assist grasp synthesis and optimization, they rely on the accurate object shapes and design complex calculations for prior integration. This makes it difficult to apply them in the 6-DoF grasp detection where the generation of diverse grasps in complex scenarios is required and the object geometry from the depth sensor is inaccurate.

\section{Method}

\begin{figure*}[t]
\centering
\includegraphics[width=0.85\linewidth]{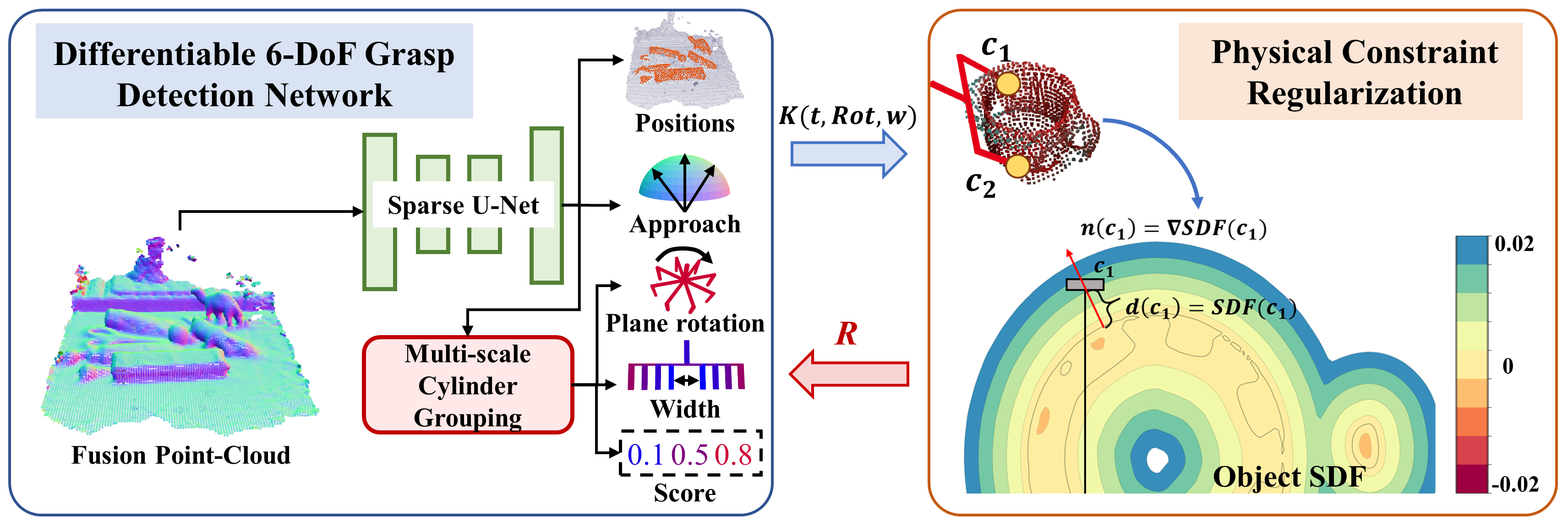}
\caption{The pipeline of differential physical constraint integration. With the fusion point-cloud, the differentiable 6-DoF grasp network predicts the grasp configurations. The position of contacts $c_1,c_2$ are calculated from grasp configurations by the gripper model and the regularization $R$ is computed from the object SDF for back-propagation.}
\label{fig:pci}
\vspace{-0.4cm}
\end{figure*}

\subsection{Physical Constraint Regularization} \label{pgp}
In terms of physical prior knowledge, the force interaction between the object and the gripper is used to analyze grasp stability, among which the force closure \cite{DBLP:journals/ijrr/Nguyen88} is widely adopted. A force closure grasp can resist any external wrenches with the contact force if the force direction lies in the friction cone. In scenarios where the friction coefficient of the object surface is unknown and a two-finger gripper is utilized, we employ the antipodal rule \cite{DBLP:journals/trob/ChenB93} as a simplification. Given two contact points $(u_1,u_2)$ lying on the object surface, the antipodal rule is formulated as:
\begin{equation}
     \left[p(u_1)-p(u_2)\right] \cdot t(u_1) = 0
\end{equation}
\vspace{-0.3cm}
\begin{equation}
     \left[p(u_2)-p(u_1)\right] \cdot t(u_2) = 0
\end{equation}
\vspace{-0.3cm}
\begin{equation}
     n(u_1) + n(u_2) = 0
\end{equation}
where $p(u)$ represents the contact position, $t(u)$ represents the unit tangent vector and $n(u)$ represents the unit outward normal vector. If the contact points $(u_1,u_2)$ between the gripper and the object comply with the antipodal rule, there is a higher probability of a successful grasp. Based on which, we introduce the PCR to constrain the output of the network with the rule.

With the predicted grasp pose $g = [t,R,w]$ and the gripper kinematic model $K$, the gripper contacts $(c_1,c_2)$ can be expressed as:
\begin{equation}
     (c_1,c_2) = K(t,R,w)
\end{equation}
To enforce the gripper contacts to comply with the antipodal constraint, we propose an antipodal regularization term, which is calculated as follows:
\begin{equation}
     R^{A}(c_1,c_2) = 1-0.5*(cos(\overrightarrow{c_1 c_2}, n(c_2)) + cos(\overrightarrow{c_2 c_1}, n(c_1)))
\end{equation}
where $\overrightarrow{c_1 c_2}$ is the vector connecting the gripper's contact points $c_1$ and $c_2$, $n(c_1)$ and $n(c_2)$ are the normal vectors at the respective contact points and $cos$ represents the cosine similarity.
Nevertheless, the predicted gripper contacts $c_1,c_2$ may not lie precisely on the object surface. To overcome this, we also introduce constraints on the distance between the gripper contacts and the object's surface. These constraints help to avoid collision and ensure that the contact points are sufficiently close to the object's surface. The collision constraint $R^C$ and surface constraint $R^S$ are formulated as:
\begin{equation}
     R^{C}(c_1,c_2) = Max(0,\theta-d(c_1))+Max(0,\theta-d(c_2))
\end{equation}
\vspace{-0.4cm}
\begin{equation}
     R^{S}(c_1,c_2) = Max(0,d(c_1)-\mu)+Max(0,d(c_2)-\mu)
\end{equation}
where the contact distance $d(c)$ is constrained between $[\theta,\mu]$. Besides, we use the grasp score predicted by the grasp detection network to weight the regularization of different grasps, thereby ensuring the coherence between the grasp score and the physical constraints. With the grasp score of $i$th grasp $s_i$, the overall physical constraint $R_i$ can be represented as:
\begin{equation}
    R_i = \frac{s_i * (R_i^{A} + R_i^{C} + R_i^{S})}{\frac{1}{M}\sum\limits_{j}\limits^{M} s_j}
\end{equation}
where $M$ is the number of seed points. The overall loss $L$ can be formulated as:
\begin{equation}
    L = L_{grasp} + \phi * R
\end{equation}
For more details of the grasp configuration loss $L_{grasp}$, please refer to the supplementary. 

\subsection{Differential Physical Constraint Integration}
To integrate the PCR into the grasp detection network, the sub-gradients of normal vector $\frac{\partial n(c)}{\partial c}$ and surface distance $\frac{\partial d(c)}{\partial c}$ with respect to the contact point $c$ should be computed. Besides, the generation of grasp poses should be differential. As a result, we introduce an end-to-end 6-DoF grasp network and the differentiable Signed Distance Function (SDF) for the calculation of PCR. As shown in Fig. \ref{fig:pci}, the configurations of the gripper are regressed from the input fusion point-cloud in an end-to-end manner. With the configurations predicted by the grasp detection network, the contact point $c$ can be calculated by the gripper kinematics model. Utilizing SDF of the object, we can query the surface distance of the contact $c_1$ directly by $d(c_1) = SDF(c_1)$ and the normal vector of $c_1$ can be calculated by $n(c_1) = \nabla SDF(c_1)$. In practice, obtaining an object's SDF directly is not feasible. Consequently, we employ a 3D grid to represent the object's SDF, where each grid point records the corresponding SDF value. During training, the SDF value at any arbitrary position is computed using tri-linear interpolation, which allows for the differentiable computation of the gradients required by the physical constraints.

\subsection{Contact-Score Joint Optimization} \label{csjo}
\begin{figure}[h]
\centering
\begin{subfigure}[]{0.5\columnwidth}
\centering
\includegraphics[width=\textwidth]
{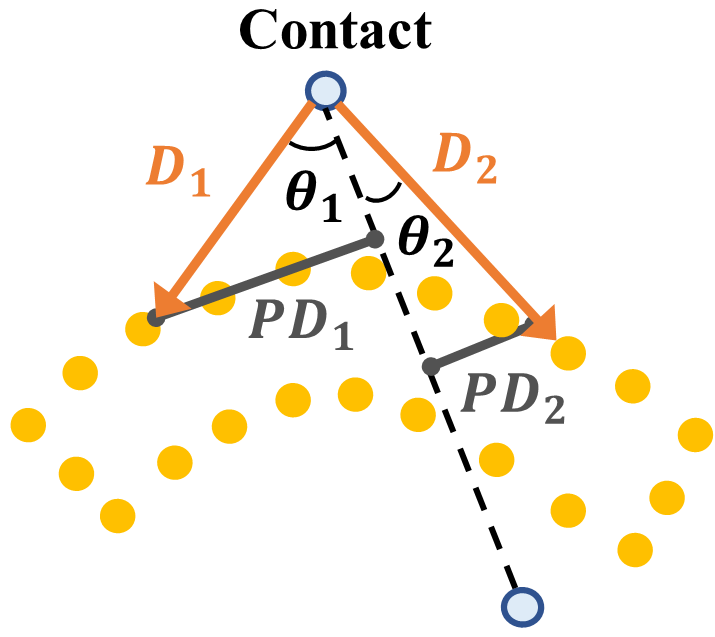}
\caption{}
\end{subfigure}
\begin{subfigure}[]{0.4\columnwidth}
\centering
\includegraphics[width=\textwidth]{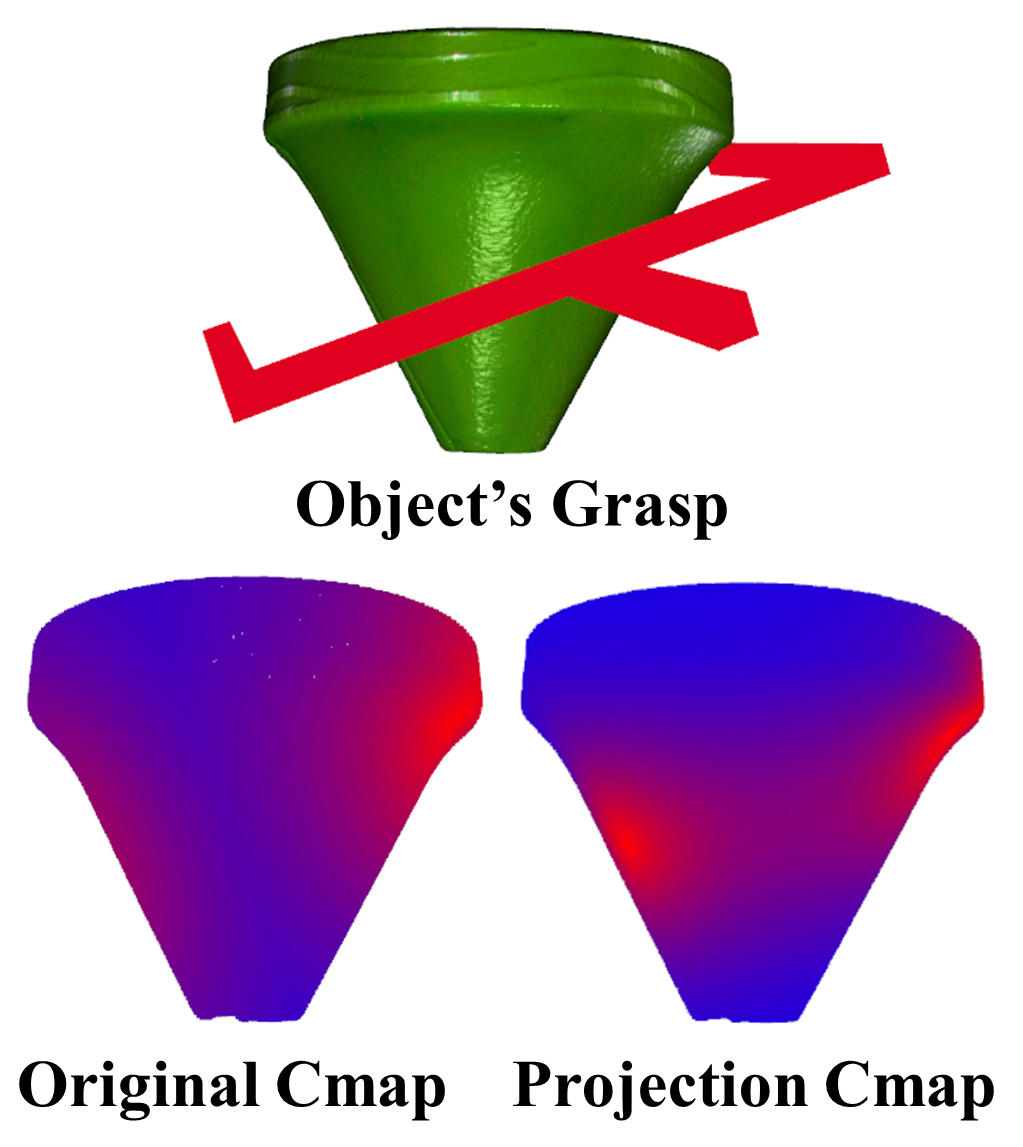}
\caption{}
\end{subfigure}
\caption{(a) Calculation of the contact map and (b) comparison of the projection contact map and the original version.}
\label{fig:contact_map}
\vspace{-0.3cm}
\end{figure}

Due to the significant discrepancy in shape and structure between novel and training objects, the grasp prediction from the network can be unstable, which leads to failures. To solve this problem, inspired by dexterous grasp synthesis \cite{DBLP:conf/iccv/JiangLW021, DBLP:conf/cvpr/XuWZLSSWGWCLYW23}, we introduce the contact map prior which encodes the preference contact region on the object to refine inaccurate grasps. Given the object point-cloud $X_o$ and gripper contacts $c_1,c_2$, we define the contact map of each point $p_i \in X_o$ for a 6-DoF grasp as:
\begin{equation}
    D_i = \min\limits_{j}\parallel p_i - c_j \parallel, c_j \in \{c_1,c_2\}
\end{equation}
Compared with optimization for a single object with accurate geometry in grasp synthesis, for the 6-DoF grasp detection task in cluttered scenes, there exist two issues that make the test-time optimization impractical and cause singular results: (1) the contact points don't always lie on the object surface; (2) the object point-clouds in real-world scenarios are inaccurate and incomplete. For the first issue, we propose the projection contact distance $PD$ illustrated in Fig. \ref{fig:contact_map} (a) as a supplementary of original contact distance $D$, which calculates the projection distance from the object point-cloud to the connecting line of two contacts. 
\begin{equation}
    PD_i = D_i * sin(\theta_i) 
\end{equation}

\begin{figure}[t]
\centering
\begin{subfigure}[]{0.9\columnwidth}
\centering
\includegraphics[width=\textwidth]
{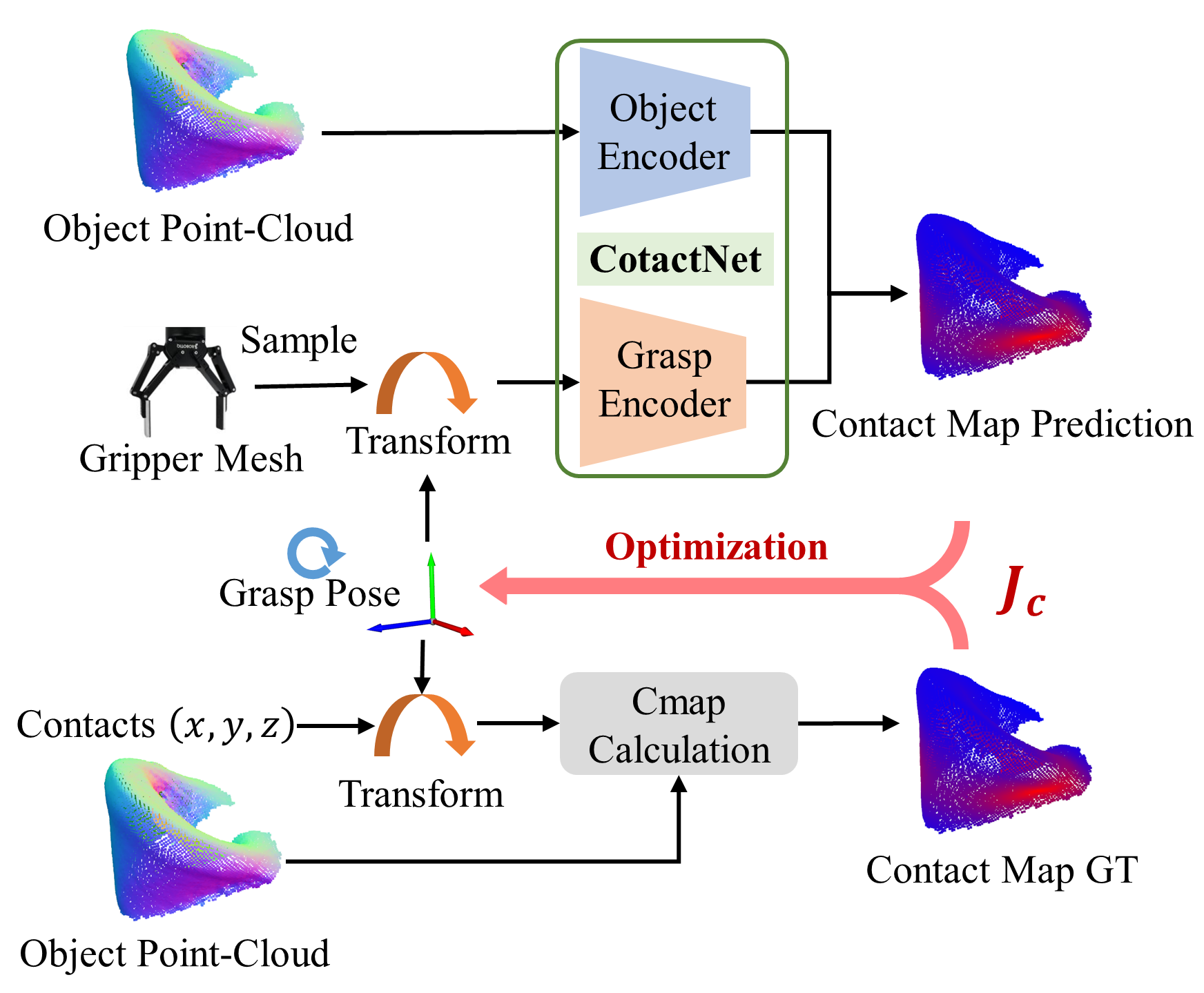}
\caption{Contact optimization}
\end{subfigure}
\begin{subfigure}[]{0.8\columnwidth}
\centering
\includegraphics[width=\textwidth]{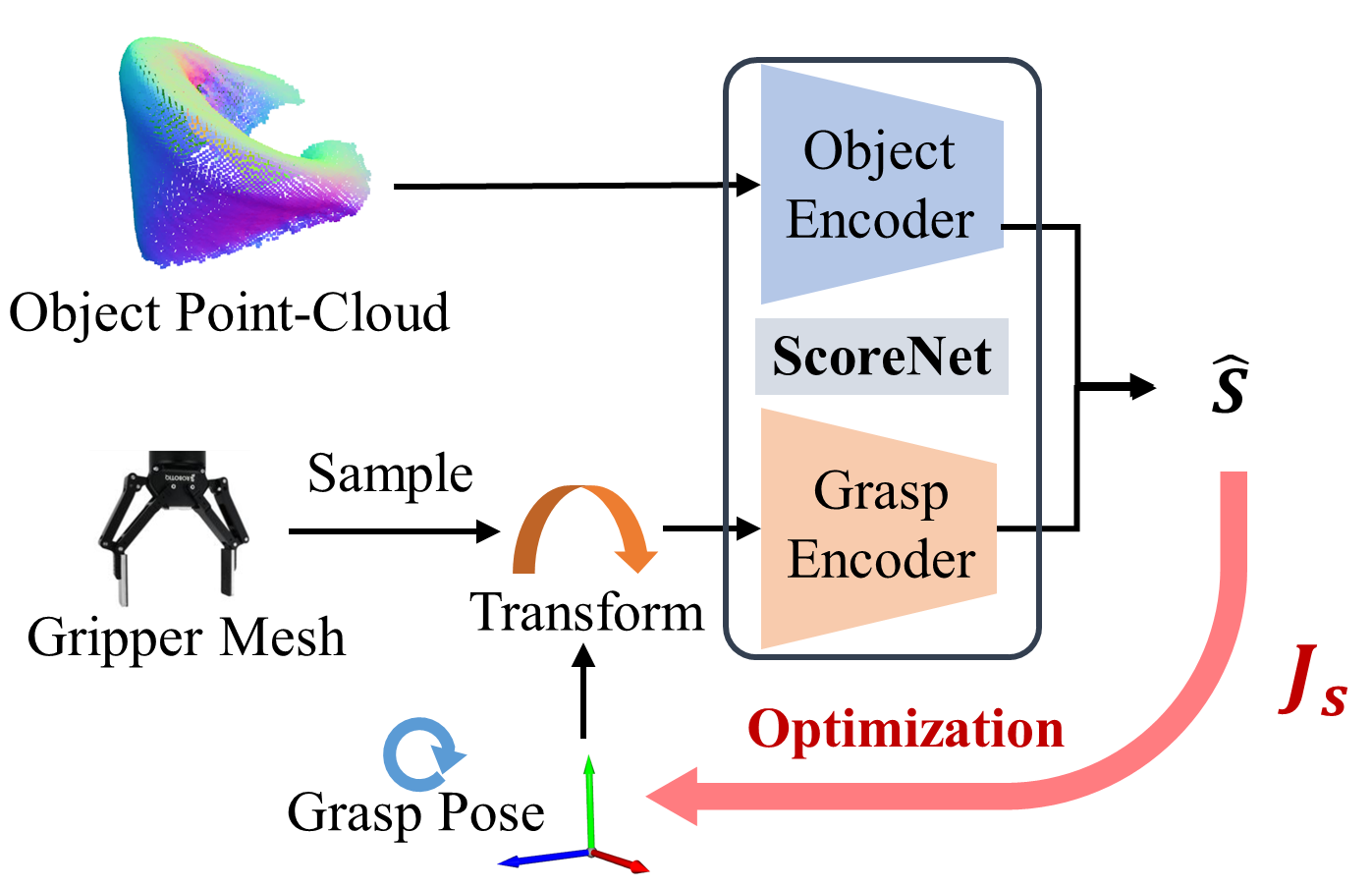}
\caption{Score optimization}
\end{subfigure}
\caption{The pipeline of contact optimization and score optimization with ContactNet and ScoreNet.}
\label{fig:csjo}
\vspace{-0.4cm}
\end{figure}

We visualize the original contact map and the proposed projection contact map in Fig. \ref{fig:contact_map} (b). Rather than computing the contact distance directly, the projection contact map can highlight the accurate contact region on the object point-cloud for coarse grasp poses whose contact points may not lie on the surface. For contact map optimization, as shown in Fig. \ref{fig:csjo} (a), we employ a ContactNet for contact map prediction, which encodes the contact prior. Given the object point-cloud $X_o$, the gripper point-cloud $X_g$ and the predicted grasp pose $g$, the ContactNet can be formulated as:
\begin{equation}
    \hat{D},\hat{PD} = ContactNet(g X_g,X_o)
\end{equation}
and thus the contact optimization target $J_c$ is formulated as:
\begin{equation}
    J_c = |D-\hat{D}| + \alpha * |PD-\hat{PD}|
\end{equation}
For the second issue, we incorporate an independent grasp score network ScoreNet in the optimization process as shown in Fig. \ref{fig:csjo} (b). Trained with the noisy object point-cloud and its corresponding grasp pose sampled from grasp labels, the ScoreNet predicts the grasp score $\hat{S}$. We sample both the good and bad quality grasps so the ScoreNet can give a low score when the failed grasp appears in the optimization. To suppress the decline in grasp score, the target of score optimization $J_s$ and the overall optimization target $J$ are formulated as:
\vspace{-0.2cm}
\begin{equation}
    J_s = T-min(T,\hat{S})
\end{equation}
\vspace{-0.4cm}
\begin{equation}
    J = J_c + \beta * J_s + \gamma * \Delta t
\end{equation}
where $T$ is the max grasp score, $\beta$ is the weight of $J_s$ and $\Delta t$ constrains the offset of position $t$ during optimization. During inference, we adopt an Adam optimizer to optimize the grasp pose iteratively by minimizing $J$ in a gradient descent strategy.

\begin{table*}[ht]
\renewcommand\arraystretch{1.2}
\centering
\begin{tabular}{l|ccc|ccc|ccc} \Xhline{1.3pt}
\multirow{2}*{\textbf{Model}} &\multicolumn{3}{c|}{\textbf{Seen}} &\multicolumn{3}{c|}{\textbf{Similar}} &\multicolumn{3}{c}{\textbf{Novel}}\\ \cline{2-10}
  & \textbf{AP} & \textbf{AP$_{0.8}$} & \textbf{AP$_{0.4}$} & \textbf{AP} & \textbf{AP$_{0.8}$} & \textbf{AP$_{0.4}$} &\textbf{AP} & \textbf{AP$_{0.8}$} & \textbf{AP$_{0.4}$} \\ \hline
Baseline & \underline{66.09} & \underline{75.57} & 60.63 & 64.82 & 74.10 & \underline{61.51}  & 30.61 & 37.61 & 17.06 \\ 
+ PCR & \textbf{66.67}& \textbf{75.67} & \textbf{61.84} & \textbf{65.55} & \textbf{74.42} & 61.47 & \textbf{35.58} & \textbf{43.49} & \underline{19.70} \\ \hline
w/o $R^{A}$ & 65.63 & 74.59 & \underline{61.19} & \underline{65.38} & \underline{74.29} & \textbf{62.10} & 32.38 & 39.04 & 18.88 \\ 
w/o $R^{C}$ & 65.27 & 75.05 & 58.62 &62.25 & 71.65 & 57.11 & 32.34 & 38.58 & 19.27 \\ 
w/o $R^{S}$ & 63.51 & 72.37 & 60.26 & 64.31 & 73.00 & 61.06 & \underline{35.57} & \underline{42.60} & \textbf{20.45}\\ \Xhline{1.3pt}
\end{tabular}
\caption{Results of PCR on scenes captured by RealSense.}
\label{table:pcr}
\end{table*}

\section{Experiments}

\subsection{Implementation Details}

\noindent \textbf{Benchmark} We conduct all the simulation experiments on the large-scale GraspNet-1billion benchmark \cite{DBLP:conf/cvpr/FangWGL20}. The benchmark includes 190 cluttered scenes, in which 100 scenes for training, 90 for testing. The testing set is divided into seen, similar and novel set based on the objects in the scenes. Each scene includes 256 RGB-D images captured from different views with RealSense/Kinect cameras. Most previous methods on the benchmark utilize single-view data for training and evaluation. However, the depth images captured from a single view suffer from significant noise. For novel object grasping, as the model has no prior about the objects, it is unable to infer the shape of occluded parts based solely on partial point-clouds, thereby hindering the grasp detection for novel objects. To bypass the interference caused by incomplete point clouds and enhance the capability of grasp detection on novel objects, we reconstruct the depth images from multiple views in a single scene into Truncated Signed Distance Function (TSDF) via KinectFusion \cite{DBLP:conf/ismar/NewcombeIHMKDKSHF11}. The TSDFs are used for training and evaluation.

\noindent \textbf{Metric} We follow the metric used in GraspNet-1billion benchmark \cite{DBLP:conf/cvpr/FangWGL20}, in which the average precision of top-\textit{k} ranked grasps in a scene is considered. In the original metric, \textit{k} is set to 50, and a maximum of 10 grasps per-object are used for evaluation. Given that most scenes contain around 9 objects, this metric can overlook the grasping accuracy of some objects. Therefore, we employ an object-balanced metric for reconstructed scenes, where $k=N_{object}*5$. $N_{object}$ is the number of objects of the scene and a maximum of 5 grasps per-object are used for evaluation. This provides a more balanced consideration of the grasping accuracy across different objects compared to the original evaluation metrics. \textbf{AP$_{\mu}$} is employed as the metric, which represents the average \textit{Precision@m} for \textit{m} ranges from 1 to \textit{m} with friction $\mu$. \textbf{AP} is calculated by the average of \textbf{AP$_{\mu}$}, where $\mu$ ranges from 0.2 to 1.2 with the interval $\Delta \mu = 0.2$.

\noindent \textbf{Model implementations} We first introduce a strong baseline model based on the scale balanced 6-DoF grasp detection network \cite{DBLP:conf/corl/Ma022} and GSNet \cite{DBLP:conf/iccv/WangFGFGL21}. Following \cite{DBLP:conf/iccv/WangFGFGL21}, we employ a sparse UNet based on Minkowski Engine \cite{choy20194d} as the backbone and calculate the "graspness" for each point to generate grasp candidates. For local feature extraction, the multi-scale cylinder grouping proposed in \cite{DBLP:conf/corl/Ma022} is employed to improve the performance of the baseline model. For the PCR during training, $\theta$ and $\mu$ which control the contact distance are set to 0.02 and 0.005 separately and the weight $\phi$ for the regularization is set to 0.1. For the C-SJO, we set the hyper-parameter $\alpha = 0.2$, $\beta = 0.01$ and $\gamma = 5$.

\subsection{Results on Physical Constraint Regularization}
We show the results of PCR in Table \ref{table:pcr}. Compared to the baseline model, integrating the PCR delivers an improvement of 4.64\% on the novel set, demonstrating its effectiveness for objects with diverse shapes and structures. Besides, the PCR also improves by 0.58\% in seen set and  0.73\% in similar set. This illustrates that for seen and similar objects in the training set, introducing physical prior can help the grasp detection network to fully utilize the grasp labels compared to fitting them directly. We also ablate the influence of different physical constraints and notice that the incomplete constraint conditions can attribute to the performance decline on seen and similar set. Only with partial constraints during training, not all the grasp poses which meet the constraints are correct, thus disturbing the learning from grasping label.

\begin{table}[h]
\renewcommand\arraystretch{1.2}
\centering
\begin{tabular}{l|c|c|c} \Xhline{1.3pt}
\textbf{Model} &\textbf{Seen} &\textbf{Similar} &\textbf{Novel}\\ \hline
No Refine & \underline{66.67} & \underline{65.55} & 35.58 \\ 
+ C-S (Predicted Mask) & 66.61 & 65.37 & \textbf{36.67} \\
+ C-S (GT Mask) & \textbf{66.69}& 65.50 & \underline{36.61} \\ \hline
Original C-S & 66.48 & 65.29 & 36.27 \\
w/o C & 66.65 & \textbf{65.64} & 35.86 \\ 
w/o S & 66.54 & 65.37 & 36.23 \\ \Xhline{1.3pt}
\end{tabular}
\caption{Results of C-SJO on scenes captured by RealSense.}
\label{table:csjoab}
\vspace{-0.3cm}
\end{table}

\begin{table*}[t]
\renewcommand\arraystretch{1.2}
\centering
\begin{tabular}{l|ccc|ccc|ccc} \Xhline{1.3pt}
\multirow{2}*{\textbf{Model}} &\multicolumn{3}{c|}{\textbf{Seen}} &\multicolumn{3}{c|}{\textbf{Similar}} &\multicolumn{3}{c}{\textbf{Novel}}\\ \cline{2-10}
  & \textbf{AP} & \textbf{AP$_{0.8}$} & \textbf{AP$_{0.4}$} & \textbf{AP} & \textbf{AP$_{0.8}$} & \textbf{AP$_{0.4}$} &\textbf{AP} & \textbf{AP$_{0.8}$} & \textbf{AP$_{0.4}$} \\ \hline
GraspNet-baseline \cite{DBLP:conf/cvpr/FangWGL20} & 28.10 & 30.53 & 26.56 & 23.87 & 26.92 & 22.51 & 7.48 & 8.43 & 4.93 \\ 
Scale-balanced Grasp \cite{DBLP:conf/corl/Ma022} & 46.05 & 51.02 & 44.27 & 37.76 & 44.27 & 34.75 & 17.09 & 21.04 & 10.20\\ 
GSNet \cite{DBLP:conf/iccv/WangFGFGL21} & 60.47 & 71.05 & 52.06 & 58.55 & 69.38 & 53.08 & 28.06 & 36.19 & 14.09 \\ \hline
Ours Baseline & 66.09 & 75.57 & 60.63 & 64.82 & 74.10 & \textbf{61.51}  & 30.61 & 37.61 & 17.06 \\ 
Ours & \textbf{66.61} & \textbf{75.67} & \textbf{61.52} & \textbf{65.37} & \textbf{74.35} & 61.28 &  \textbf{36.67} & \textbf{45.08} & \textbf{20.90} \\ \hline \Xhline{1.3pt}
\end{tabular}
\caption{Comparison with the state-of-the-art methods on RealSense scenes of GraspNet-billion benchmark.}
\label{table:sota}
\end{table*}

\begin{figure*}[h]
\centering
\includegraphics[width=0.85\linewidth]{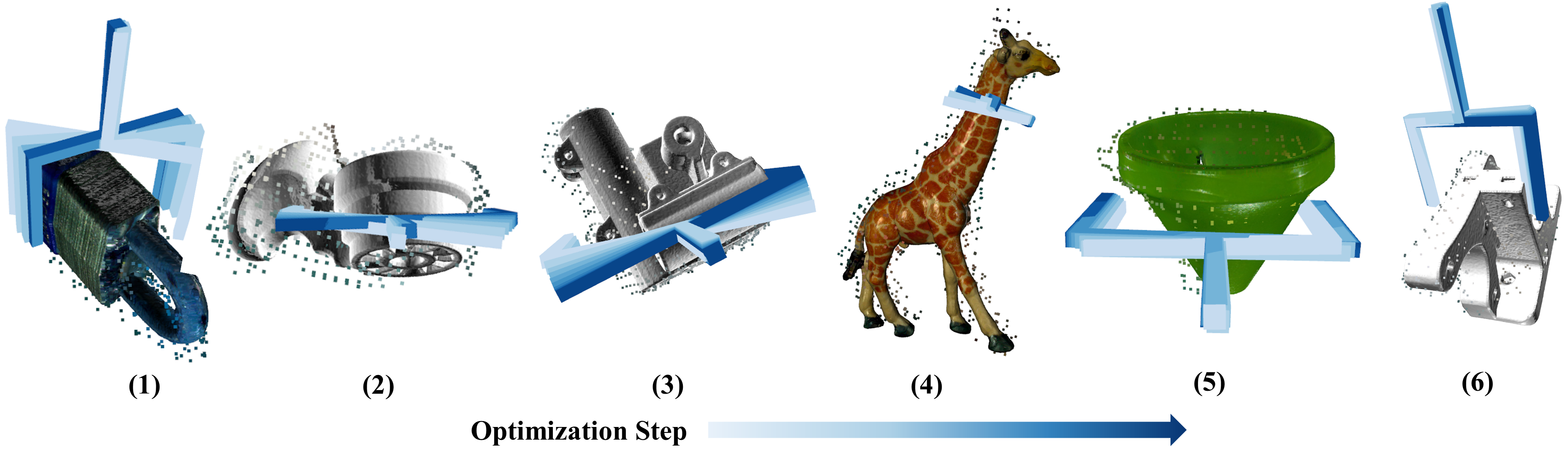}
\caption{Visualization of the process of C-SJO.}
\label{fig:visopt}
\vspace{-0.2cm}
\end{figure*}

\subsection{Results on Contact-Score Joint Optimization}

We employ the C-SJO in our 6-DoF grasp detection network trained with PCR and conduct ablations. The results are shown in Table \ref{table:csjoab}. We incorporate a recent proposed 3D segmentation method \cite{yang2023sam3d} to get object point-clouds from the clutter for optimization. With the ground-truth mask and segmented mask, C-SJO can improve by 1.03\% and 1.00\% on novel set separately. The C-SJO has almost no impact on the seen and similar set, primarily because the grasp poses predicted for seen and similar objects are sufficiently stable, making it challenging to find a better grasp than the original one. However, for the novel set, many unstable grasps exist that can potentially be optimized. We also conduct ablations about different designs of the C-SJO. Without the proposed projection contact map, only employing the original contact map (Original C-S) drops on seen, similar and novel set due to the singularity poses during optimization. Optimizing solely based on grasp score (w/o C) overlooks the contact map prior, resulting in very limited improvement on the novel set. Using only the contact map for optimization (w/o S) yields a 0.38\% drop on the novel set, but the absence of grasp scores restricts the search space, making it challenging to locate the locally optimal grasp during the optimization process. The process of grasp pose optimization is shown in Fig. \ref{fig:visopt}, where the optimal grasp poses are reached in the original grasps' neighbor $SE(3)$ space. For object (1)-(4), the optimization process refines the inaccurate initial grasp poses. For objects (5) and (6), the optimization process facilitates collision avoidance.

\subsection{Comparison with State-of-the-art}

To make a fair comparison with the state-of-the-art methods of the GraspNet-1billion benchmark, we re-implement three representative 6-DoF grasp detection methods \cite{DBLP:conf/cvpr/FangWGL20, DBLP:conf/corl/Ma022, DBLP:conf/iccv/WangFGFGL21} with our multi-view reconstructed scene as input. As shown in Table \ref{table:sota}, our method performs better than previous 6-DoF grasp detection methods on all test sets and achieves 36.67\% on the novel set, which demonstrates the generalization capability of our method.

\subsection{Comparison with object augmentation}

We give a comparison between our prior knowledge based method and the object augmentation which employed by previous methods to enhance the generalization capability. To migrate the procedural or learning-based object augmentation method on the GraspNet-1billion benchmark, since the similar objects and the training objects have similar distributions, we consider the similar set in the benchmark as objects obtained through the augmentation. We merge the two sets for joint training. In this way, we can conveniently validate the performance of data augmentation on this benchmark without the need to generate additional objects and annotate grasps. As shown in Table \ref{table:aug}, introducing object augmentation can slightly improve the performance on seen and novel sets, benefiting from the richer object distribution. However, our method performs better on the novel set, improving by 4.37\% compared to the object augmentation method, without introducing any additional data. Simultaneously, our method also exhibits an increase of 0.33\% on novel set when paired with object augmentation, demonstrating that our method and object augmentation can be used together.

\begin{table}[h]
\renewcommand\arraystretch{1.2}
\centering
\begin{tabular}{l|c|c|c} \Xhline{1.3pt}
\textbf{Model} &\textbf{Seen} &\textbf{Similar} &\textbf{Novel}\\ \hline
Baseline & 66.09 & 64.82 & 30.61 \\ 
Augmentation & 68.12 & - & 32.30 \\
Ours & 66.61& 65.37 & 36.67 \\
Ours + Augmentation & \textbf{68.31} & - & \textbf{37.00} \\ \Xhline{1.3pt}
\end{tabular}
\caption{Comparison with object augmentation.}
\label{table:aug}
\vspace{-0.4cm}
\end{table}

\begin{figure*}[t]
\centering
\includegraphics[width=0.85\linewidth]{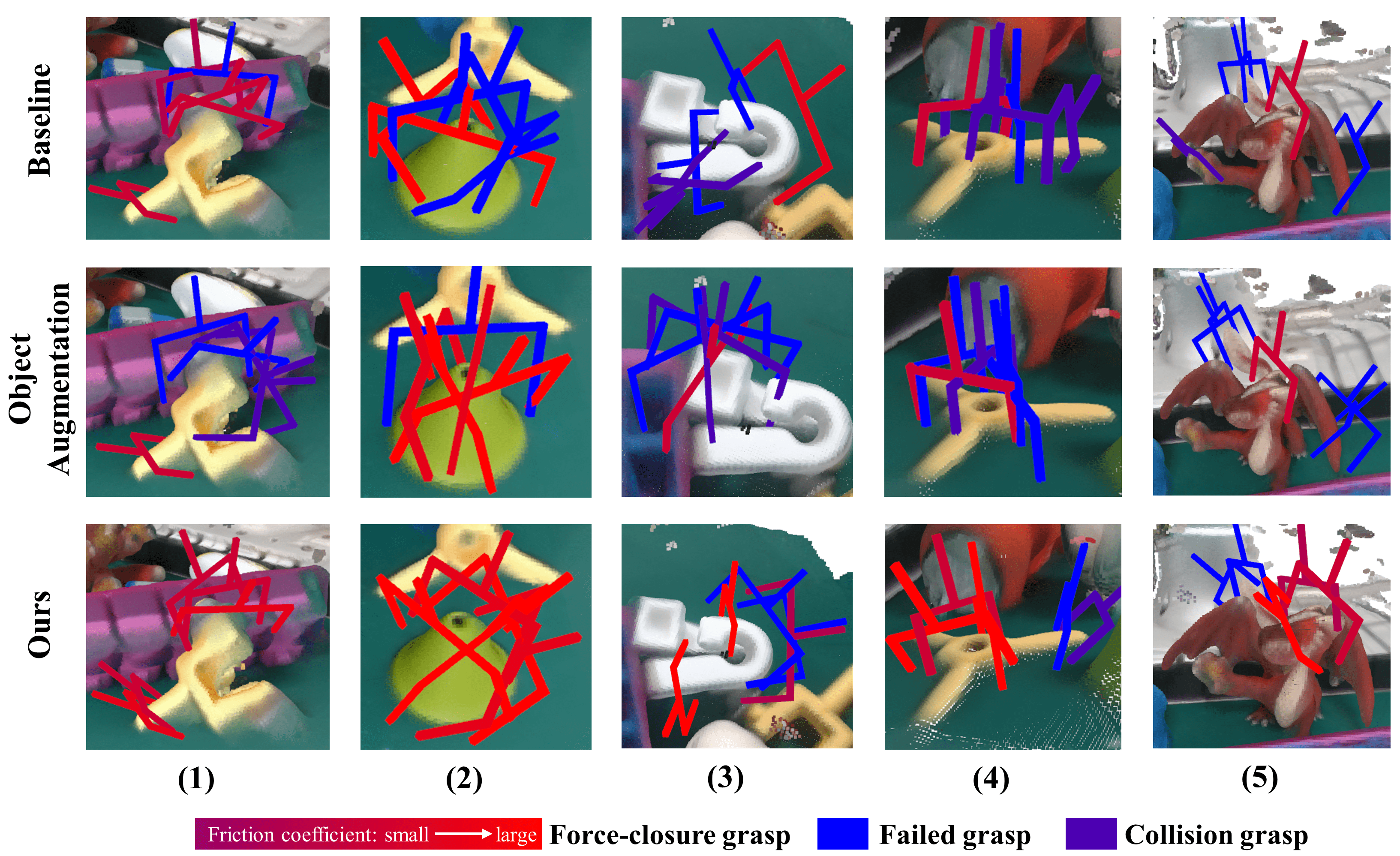}
\caption{Visualization of the predicted grasp poses from baseline, object augmentation and the methods proposed in this paper.}
\label{fig:resvis}
\vspace{-0.2cm}
\end{figure*}

\subsection{Result Visualization}

In figure \ref{fig:resvis}, we visualized the results generated by the baseline method, object augmentation and our approach on the GraspNet-1billion benchmark. The gripper poses in red represent successful grasps, while those shown in purple and blue correspond to collision and bad grasps. By incorporating domain prior knowledge, our method can generate grasp poses that conform to the physical relationship between the gripper and the object. For objects (3)-(5), our method predicts some failed grasps, primarily due to poor grasp sampling locations or inaccurate perception of the object's geometry.

\subsection{Real-world Evaluation}

\begin{figure}[h]
\centering
\begin{subfigure}[]{0.43\columnwidth}
\centering
\includegraphics[width=\textwidth]
{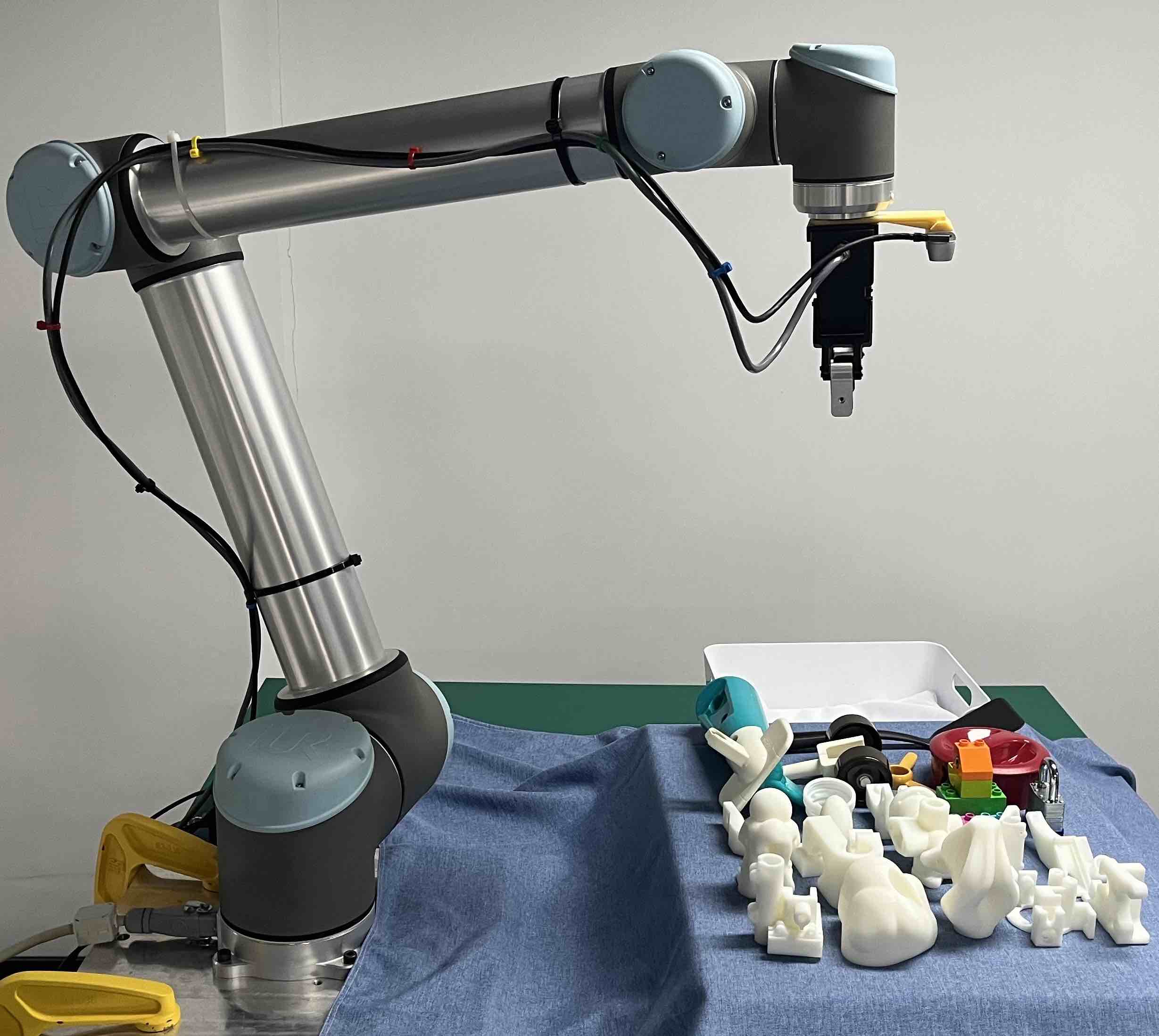}
\end{subfigure}
\begin{subfigure}[]{0.5\columnwidth}
\centering
\includegraphics[width=\textwidth]{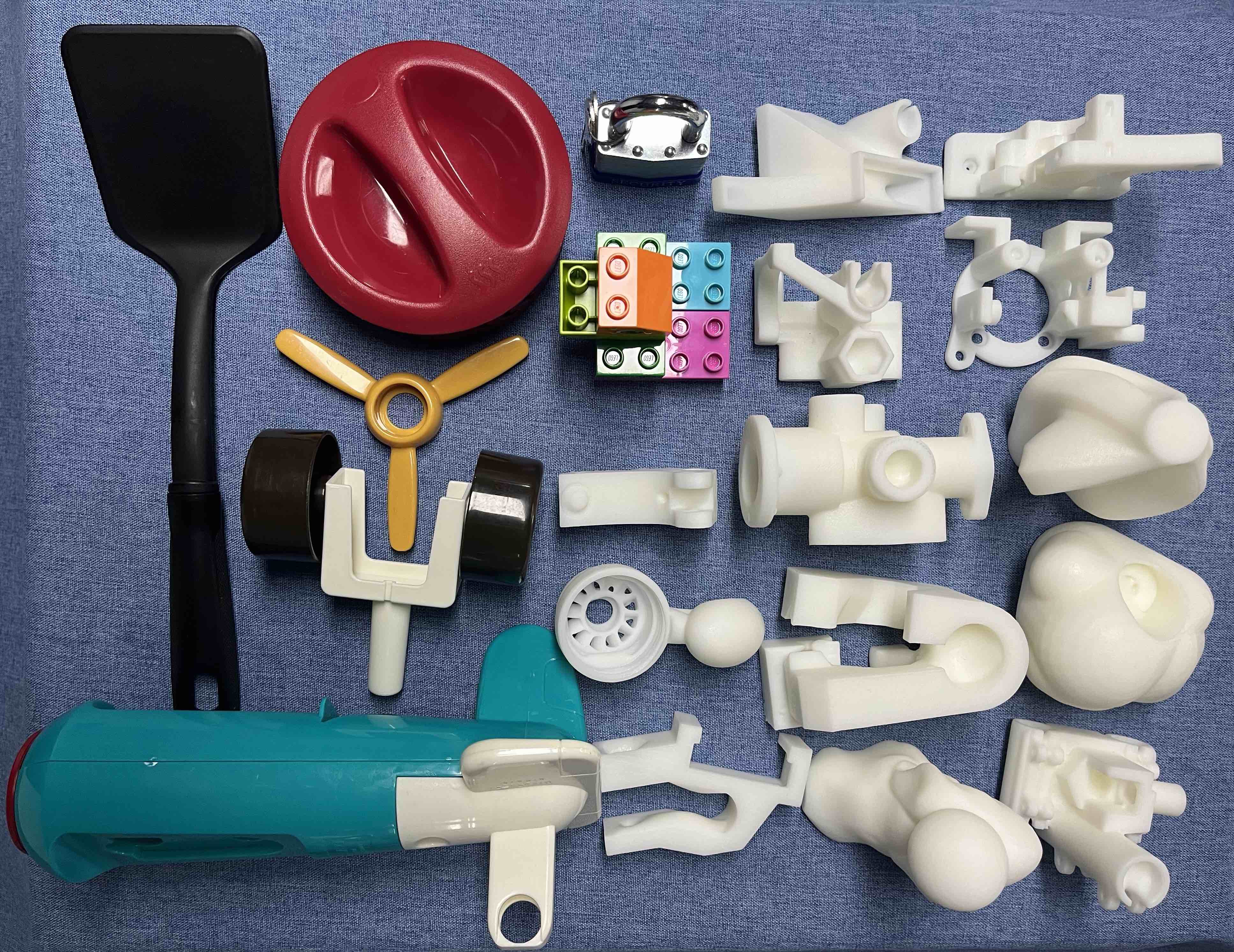}
\end{subfigure}
\caption{(a) Robotic grasping system and (b) objects for real-world grasping.}
\label{fig:real-world}
\vspace{-0.4cm}
\end{figure}

The effectiveness of the proposed method in real-world is validated by real robot in this section. As shown in Figure \ref{fig:real-world} (a), the robotic grasping system is built on a 6-DoF UR-10 robot arm and a RealSense D435i depth camera is employed for scene perception. The total of 20 objects used for grasping (Fig. \ref{fig:real-world} (b)) are composed of two parts: 13 3D-printed objects from Dex-Net \cite{DBLP:conf/rss/MahlerLNLDLOG17} and 7 objects chosen from the YCB dataset \cite{DBLP:journals/ijrr/CalliSBWKSAD17}. All of the objects are significantly different from the objects used for training in GraspNet-1billion. Before executing the proposed grasp detection method, we quickly reconstruct the scene using an arm-mounted depth camera based on the KinectFusion \cite{DBLP:conf/ismar/NewcombeIHMKDKSHF11}.

\begin{table}[h]
\centering
\small
\begin{tabular}{l|c|cc}\Xhline{1.3pt}
\multirow{2}*{\textbf{Model}} &\multicolumn{1}{c|}{\textbf{Isolated}} &\multicolumn{2}{c}{\textbf{Cluttered}} \\ \cline{2-4} & \textbf{SR ($\%$)} & \textbf{SR ($\%$)} & \textbf{SCR ($\%$)} \\ \hline
Baseline & 58.33 (35/60) & 48.21 (43/89) & 86.00 (43/50) \\ \hline
Ours & 68.33 (41/60) & 64.86 (48/74) & 96.00 (48/50) \\ \Xhline{1.3pt}
\end{tabular}
\caption{Results of the real-world grasping experiments.}
\label{table:real-world}
\vspace{-0.3cm}
\end{table}

We compare our model to the baseline in two settings: isolated object grasping and cluttered object grasping. For isolated object grasping, each object is placed in three different poses and Success Rate (SR) is used as the metric. For cluttered object grasping, we compose 5 objects into a scene and make the robot remove them all with a maximum number of operations at 10. SR and Scene Completion Rate (SCR) are employed as the metrics. As illustrated in Table \ref{table:real-world}, our model outperforms the baseline in both settings and achieves better SR and SCR, demonstrating its superiority for novel objects.

\section{Conclusion}
In this paper, we work for generalizing 6-DoF grasp detection with domain prior knowledge of robotic grasping. The physical constraint regularization based on physical rules is proposed to enable the generalization of objects with largely varied shapes and structures.  To refine the unstable results predicted by network in cluttered scenarios, we specially design a contact-score joint optimization with the contact map prior, in which a projection contact map is utilized. Extensive experiments on both the benchmark and the real-world robot demonstrate the effectiveness of our method for novel objects.

\section*{Acknowledgment} 
This work is partly supported by the National Key R\&D Program of China (2022ZD0161902), the National Natural Science Foundation of China (62022011), the Research Program of State Key Laboratory of Software Development Environment (SKLSDE-2023ZX-14), and the Fundamental Research Funds for the Central Universities.

\clearpage
{\small
\bibliographystyle{ieeenat_fullname}
\bibliography{egbib}
}

\end{document}